\documentclass[10pt,twocolumn,letterpaper]{article}

\usepackage{ijcb}
\usepackage{times}
\usepackage{epsfig}
\usepackage{graphicx}
\usepackage{amsmath}
\usepackage{amssymb}
\usepackage[norule,symbol,perpage]{footmisc}
\usepackage{multirow}
\usepackage{subcaption}

\graphicspath {{figures/}}
\newcommand{\ppname}{DBLFace\xspace}

\newcommand{\norm}[1]{\left\lVert#1\right\rVert}

\usepackage[pagebackref=true,breaklinks=true,letterpaper=true,colorlinks,bookmarks=false]{hyperref}

\ijcbfinalcopy %

\ifijcbfinal\fi

\makeatletter
\def\ps@IEEEtitlepagestyle{
\def\@oddfoot{\mycopyrightnotice}
\def\@evenfoot{}
}
\def\mycopyrightnotice{
{\hfill \footnotesize 978-1-7281-9186-7/20/\$31.00 \copyright 2020 IEEE\hfill}
}
\makeatother

\begin{document}

\title{\ppname: Domain-Based Labels for NIR-VIS Heterogeneous Face Recognition}

\author{Ha A. Le and Ioannis A. Kakadiaris\\
Computational Biomedicine Lab\\
Dept. of Computer Science, University of Houston\\
{\tt\small \{hale4, ikakadia\}@central.uh.edu}
}

\maketitle
\thispagestyle{empty}

\begin{abstract}
   Deep learning-based domain-invariant feature learning methods are advancing in near-infrared and visible (NIR-VIS) heterogeneous face recognition. However, these methods are prone to overfitting due to the large intra-class variation and the lack of NIR images for training. In this paper, we introduce Domain-Based Label Face (\ppname), a learning approach based on the assumption that a subject is not represented by a single label but by a set of labels. Each label represents images of a specific domain. In particular, a set of two labels per subject, one for the NIR images and one for the VIS images, are used for training a NIR-VIS face recognition model. The classification of images into different domains reduces the intra-class variation and lessens the negative impact of data imbalance in training. To train a network with sets of labels, we introduce a domain-based angular margin loss and a maximum angular loss to maintain the inter-class discrepancy and to enforce the close relationship of labels in a set. Quantitative experiments confirm that \ppname significantly improves the rank-1 identification rate by 6.7\% on the EDGE20 dataset and achieves state-of-the-art performance on the CASIA NIR-VIS 2.0 dataset.
\end{abstract}

\section{Introduction} \label{sec:introduction}
NIR-VIS heterogeneous face recognition refers to the problem of matching face images across the two visual domains. It has been widely adopted to various applications, such as video surveillance and user authentication in deficient lighting conditions. With the evolution of deep learning models, a number of deep learning-based methods \cite{Fu2019, Deng2019, Hu2019} are presented and achieved significant improvement on the popular benchmarks (\eg, CASIA NIR-VIS 2.0 \cite{Li2013} and Oulu-CASIA NIR-VIS \cite{JieChen2010}). However, NIR-VIS heterogeneous face recognition remains a challenging problem due to the large cross-modality gap and the lack of large-scale training data with both VIS and NIR images.

\begin{figure}[t]
	\begin{center}
		\begin{subfigure}[b]{0.55\columnwidth}
			\includegraphics[width=\textwidth]{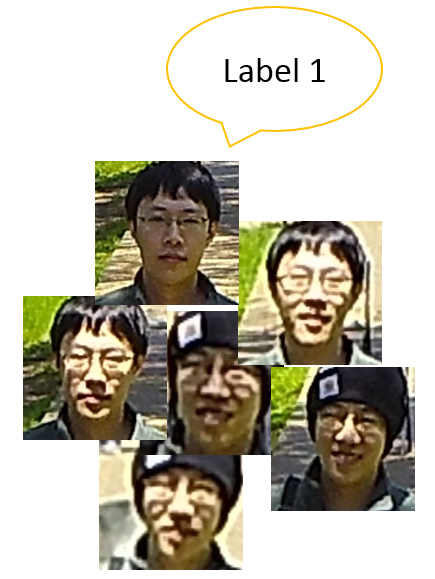}
			\caption{}
			\label{fig:vislabel}
		\end{subfigure}
		\begin{subfigure}[b]{0.35\columnwidth}
			\includegraphics[width=\textwidth]{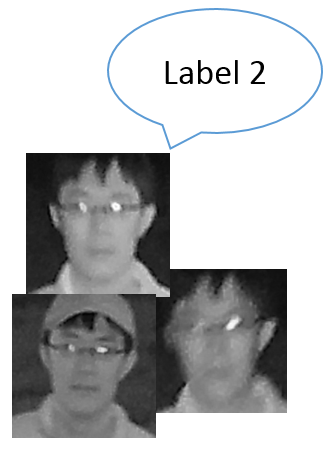}
			\caption{}
			\label{fig:nirlabel}
		\end{subfigure}
	\end{center}
	\caption{Depiction of a set of two labels of a subject: (a) the VIS images with label 1 and (b) the NIR images with label 2. The images depicted on (a) and (b) belong to the same subject, but they are labeled as two separate classes.}
	\label{fig:labels}
\end{figure}

Datasets of VIS images (\eg, MS1M \cite{Guo2016} and Glint \cite{DeepGlint2018}) that contain millions of faces with various face poses and illuminations play a vital role in the success of face recognition algorithms in the VIS domain. In comparison to these datasets, cross-domain face datasets contain a significantly fewer number of subjects and images. For instance, CASIA NIR-VIS 2.0, one of the largest cross-spectral face datasets, contains only 17,580 images of 725 subjects captured in a constrained environment. The amount of data from CASIA NIR-VIS 2.0 is insufficient for training a face recognition system that can accurately identify images in both visual domains. Several deep learning-based domain-invariant feature learning methods \cite{Juefei-Xu2015a, He2018, Hu2019} have been proposed to overcome the high intra-class variation by training on a large-scale dataset of VIS images and fine-tuning on a small-scale dataset of both VIS and NIR images. Although training or fine-tuning on a small-scale dataset alleviates the gap between the source and the target domains, it also reduces the generalization capability of the trained models. The lack of generalization leads to poor performance on new data with much difference to the training data.

\begin{figure*}[t]
	\begin{center}
		\begin{subfigure}[b]{0.45\textwidth}
			\includegraphics[width=\textwidth]{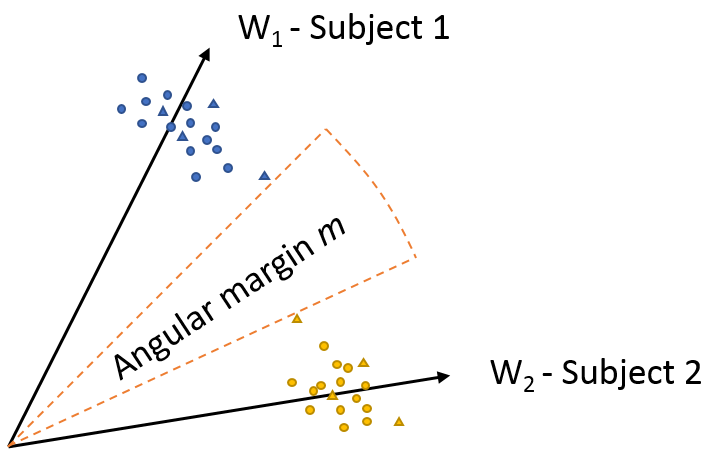}
			\caption{ArcFace}
			\label{fig:arcface}
		\end{subfigure}	
		\begin{subfigure}[b]{0.45\textwidth}
			\includegraphics[width=\textwidth]{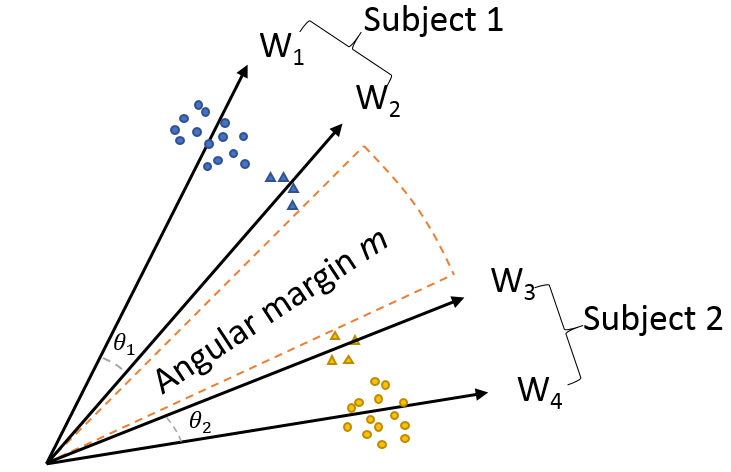}
			\caption{DBLFace}
			\label{fig:dblface}
		\end{subfigure}
	\end{center}
	\caption{Geometrical interpretation of \ppname in comparison with ArcFace. The blue and yellow points represent the features of two different subjects. The points denoted by a circle represent the VIS images, and the points denoted by a triangle represent the NIR images. There are more circles than triangles since there are more VIS images than NIR images. In ArcFace, $W_1$ and $W_2$ represent two different class representations of two subjects. In \ppname, $W_1$ and $W_2$ are the class representations of the VIS and NIR labels of Subject 1. Similarly, $W_3$ and $W_4$ are the class representations of the NIR and VIS labels of Subject 2. In ArcFace, the inter-class discrepancy is enforced by the angular margin $m$. In \ppname, the inter-class discrepancy is enforced only between class representations of different subjects. The relationship between $W_1$ and $W_2$ is enforced by minimizing the angle $\theta_1$, and the relationship between $W_3$ and $W_4$ is enforced by minimizing the angle $\theta_2$.}
	\label{fig:arcface_vs_dbface}
\end{figure*}

To avoid fine-tuning on a small-scale dataset, we propose to train a face recognition model on a joint dataset of a large-scale dataset of VIS images and a small-scale dataset of both VIS and NIR images. The number of NIR images in the joint dataset is significantly fewer than the number of VIS images. Therefore, the NIR images become outliers in their classes. To enhance the contribution of NIR images to the training process, we assume that a subject is not represented by a single label but by a set of labels. Each label represents images of a specific visual domain. In particular, a set of two labels per subject, one for the NIR images and one for the VIS images, are used for training a NIR-VIS face recognition model. Based on this assumption, NIR images can contribute to the learning of their class representations without being dominated by VIS images. In addition, the classification of images of each subject into two different labels reduces the intra-class variation since each class contains images of only one visual domain.

Inspired by the additive angular margin loss (ArcFace) \cite{Deng20181}, we introduce a domain-based angular margin loss (DAML) to maintain the inter-class discrepancy. Based on our assumption, the relationship between classes is not equal. It is redundant to enforce the discrepancy between classes that correspond to the same subject. Therefore, the DAML is designed to enforce the margin between classes of different subjects only. In addition, the classes that correspond to the same subject should be close to each other in the representation space. Therefore, we introduce a maximum angular loss (MAL) to minimize the angle between class representations that correspond to the same subject. The geometrical interpretation of our DAML and MAL is depicted in Fig. \ref{fig:arcface_vs_dbface}.

The key contributions of this work are as follows:
\begin{itemize}
	\item We introduce \ppname, a learning approach based on the assumption that a subject is represented by a set of two labels, one for the VIS images and one for the NIR images. The classification of images reduces the intra-class variation and lessens the negative impact of data imbalance in training (Section \ref{sec:introduction}).
	\item We propose a DAML to enforce the angular margin between classes of different subjects (Section \ref{sec:daml}).
	\item We propose a MAL that controls the close relationship of class representations of the same subject. The MAL allows class representations of the same subject to be close in the representation space (Section \ref{sec:mal}).
\end{itemize}

\section{Related Work} \label{sec:relatedwork}
In this section, we summarize deep learning-based methods for NIR-VIS heterogeneous face recognition. We focus on the methods presented in the last five years. For other works, please refer to \cite{Farokhi2016, Ouyang2016}. Deep learning-based methods for NIR-VIS heterogeneous face recognition can be categorized into three groups: adversarial learning, latent subspace learning, and domain-invariant feature learning. The three categories of NIR-VIS heterogeneous face recognition methods are independent and can be combined together to improve face recognition performance. The adversarial learning methods can be used as a pre-processing step, which generates new images for training and matching. The latent subspace learning methods can be used as a post-processing step, which can enhance the feature discrimination across different domains. The domain-invariant feature learning methods directly seek discriminative features by optimizing network designs and loss functions.

\textbf{Adversarial learning.} Adversarial learning methods aim to synthesize both VIS and NIR images so that the synthesized images can be used for training or matching with the existing face recognition models. The first deep learning method that attempted to synthesize VIS images from NIR images is introduced by Lezama \etal \cite{Lezama2017}. In this work, three CNNs are trained on pairs of corresponding NIR-VIS patches. Each CNN is trained to convert an image channel in the YCbCr color space. Even though Lezama's method can produce VIS images from NIR images, the synthesized images look unrealistic. With the development of the Generative Adversarial Networks (GANs) \cite{Goodfellow2014}, several methods are able to produce high-quality VIS images. Song \etal \cite{Song2018} integrate cross-spectral face hallucination and discriminative feature learning into an end-to-end adversarial network. Song's method used a two-path model to learn from both global structures and local textures. Instead of directly generating VIS images, He \etal \cite{He2020} introduced a cross-spectral face completion method that generates a facial texture map and a UV map. The VIS image is generated by a warping network that warps the facial texture map and the UV map together. Recently, Fu \etal \cite{Fu2019} introduced the Dual Variational Generation (DVG) method that promotes the inter-class diversity by generating massive new pairs of VIS and NIR images from noise. The synthesized pairs are used in training heterogeneous face recognition models to reduce domain discrepancy.
Although adversarial learning methods can generate new images, the identity of the generated images is not well preserved. A pretrained face recognition model is used to control the identity of the generated images. Therefore the precision of the identity is limited by the performance of the pretrained model.

\textbf{Latent subspace learning.} Latent subspace learning methods project the extracted features from two different domains to a common latent subspace. The feature matching is conducted on the latent subspace. Cho \etal \cite{Cho2019} presented a post-processing relation module to capture relations and coordinates of the pairwise feature to reduce the domain discrepancy. In addition, a triplet loss with a conditional margin is introduced to reduce intra-class variation in training. Peng \etal \cite{Peng2019} proposed a locally linear re-ranking technique to refine the initial ranking results. The proposed re-ranking method does not require any human interaction or data annotation and can be categorized as an unsupervised post-processing technique. Since latent subspace learning methods are post-processing methods, they require additional processing time in face matching applications.

\textbf{Domain-invariant feature learning.} Domain-invariant feature learning methods attempt to alleviate domain information and learn discriminative features by introducing new network designs and new learning metrics. Due to the lack of training data, some methods \cite{Saxena2016, Liu2016b} chose to finetune the last one or two layers of deep face recognition models using logistic discriminant metric \cite{Guillaumin2009} or triplet loss \cite{Schroff2015}. Wu \etal \cite{Wu2018} presented a cross-modal ranking among triplet domain-specific images to maximize the margin for different identities and increase data for a small number of training samples. He \etal \cite{He2018} used a siamese network to learn a modality-invariant feature subspace by minimizing the Wasserstein distance between the distributions of NIR features and VIS features. Hu \etal \cite{Hu2019b} designed a dimension reduction block to effectively extract the auxiliary features on multiple mid-level layers and reduce spectrum variations of two different modalities. In addition, they introduced the scatter loss to embed both inter-class and intra-class information for effectively training the deep model. The DSVNs model \cite{Hu2019} presented disentangled spectrum variation blocks to remove spectrum information step-wise and learned identity discriminative features. Deng \etal \cite{Deng2019} introduced a residual compensation module to compensate for the variation of features in different domains and learn discriminative features by minimizing the modality discrepancy loss. Cao \etal \cite{Cao2019a} presented a multi-margin loss to minimize cross-domain intra-class distance and further maximize cross-domain inter-class distance. Cho \etal \cite{Cho2020} introduced a graph-structured module called Relational Graph Module (RGM) that focuses on the high-level relational information between facial components. The RGM embeds spatially correlated feature vectors into graph node vectors and performs relation modeling between them.
Domain-invariant feature learning methods are prone to overfitting because they are finetuned on a small-scale dataset of VIS and NIR images. Our proposed method belongs to the domain-invariant feature learning category, but it is trained on the joint dataset to maintain model generalization.

\begin{figure*}[t]
	\begin{center}
		\includegraphics[width=1.0\linewidth]{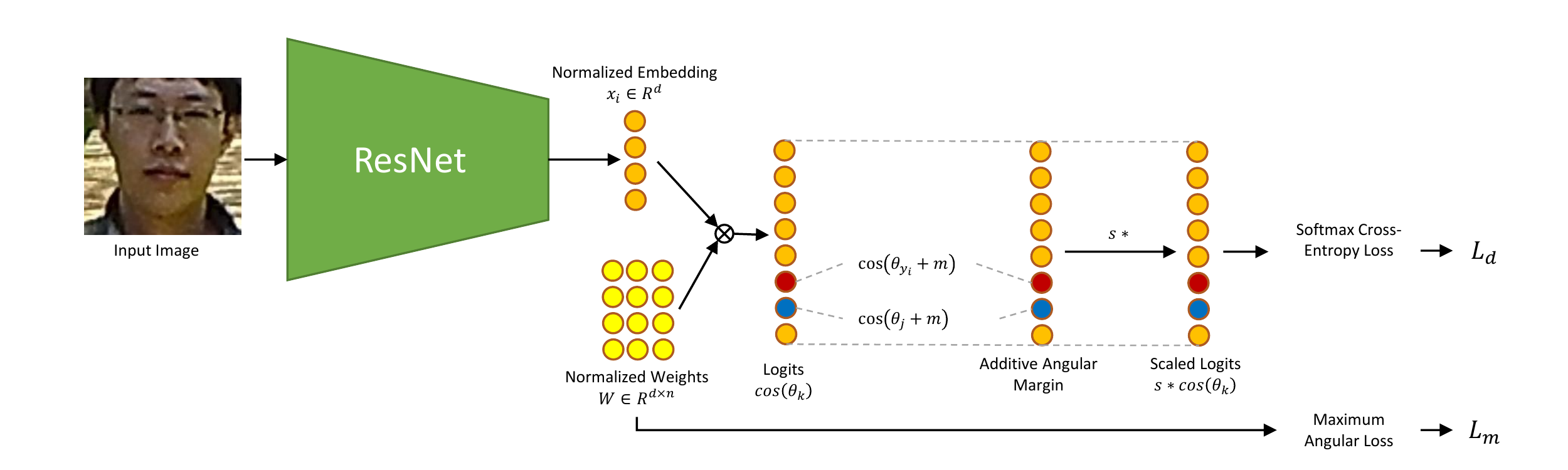}
	\end{center}
	\caption{Training procedure for \ppname. Given an input image $I$, a normalized feature embedding $x_i \in R^d$ is learned using a residual network. The normalized weights $W \in R^{d\times n}$ are presented to transform the feature embedding $x_i$ to logits. The logit  $\cos\theta_k$ for each class is computed as the dot product between the normalized feature embedding $x_i$ and the column $W_k$ of $W$. The additive angular margin $m$ is added to the corresponding logits. The red logit represents the logit of the target class $y_i$, and the blue logit represents the logit of the class has the same subject with the target class $y_i$. All logits are scaled by a constant $s$ before passing through the softmax cross-entropy loss function. The output of softmax cross-entropy loss is our DAML $L_d$. The MAL $L_m$ is computed directly from the normalized weights $W$.}
	\label{fig:dblface-network}
\end{figure*}

\section{Method} \label{sec:method}
The overall training procedure for \ppname is depicted in Fig. \ref{fig:dblface-network} with two loss functions: DAML $L_d$ and MAL $L_m$.

\subsection{Domain-based Angular Margin Loss (DAML)} \label{sec:daml}
Based on our assumption, images of a subject are split into multiple classes. Each class represents images of a specific visual domain. The DAML is designed to enforce the angular margin between classes of different subjects. The function of our DAML is derived from the softmax loss \cite{Liu2017} and the additive angular margin loss \cite{Deng20181}. Instead of only adding the margin $m$ to the logit of the target class $y_i$, the additive margin $m$ is also added to the logits of the classes that correspond to the same subject with the target class $y_i$. The addition to both the target class and the classes that correspond to the same subject with the target class eliminated the margin enforcement between these classes.

The origin softmax loss (\ie, cross-entropy loss) is formulated as:
\begin{equation}
L_s = -\frac{1}{N}\sum_{i=1}^{N}\log\frac{e^{W^T_{y_i}x_i+b_{y_i}}}{\sum_{j=1}^{n}e^{W^T_{j}x_i+b_{j}}},
\end{equation}
where $N$ and $n$ are the batch size and the number of classes, respectively, $x_i \in R^d$ denotes the $d$-dimensional feature embedding of the $i^{th}$ sample, $y_{i} \in \lbrack1,...,n\rbrack$ is the target class of $x_i$, $W_j \in R^d$ denotes the $j^{th}$ column of the weight ${W \in R^{d \times n}}$, and $b_j \in R$ is the bias of the $j^{th}$ class.

The angular margin loss assumes that all feature embeddings are distributed on a hypersphere of radius $s$. The assumption is obtained by fixing the bias $b_j=0$ and normalizing the weight $W_j$ and the feature embedding $x_i$ by $L_2$ normalization. The feature embedding $x_i$ is then re-scaled by the radius $s$ of the hypersphere so that $\norm{x_i}=s$. The additive angular margin is defined as follows:
\begin{equation}
L_a = -\frac{1}{N}\sum_{i=1}^{N}\log\frac{e^{s\left(\cos\left(\theta_{y_i}+m\right)\right)}}{e^{s\left(\cos\left(\theta_{y_i}+m\right)\right)}+\sum_{j=1,j\ne y_i}^{n}e^{s\cos\theta_j}},
\end{equation}
where $m$ is the additive margin penalty, $\theta_{y_i}$ is the angle between the weight $W_{y_i}$ and the feature embedding $x_i$, and $\theta_j$ is the angle between the weight $W_j$ and the feature embedding $x_i$. The additive margin $m$ enforces the intra-class compactness and inter-class diversity.

To eliminate the enforcement of the additive margin $m$ to the classes that correspond to the same subject with the target class $y_i$, the margin $m$ is also added to the angles between the feature embedding $x_i$ and the weights $W_{j}, j \in D_{y_i}$. The symbol $D_{y_i}$ denotes the set of classes that correspond to the same subject with the target class $y_i$. Therefore, the DAML function is formulated as follows:
\begin{equation}
\scalebox{0.75}[1]{$
	L_d = -\frac{1}{N} \sum_{i=1}^{N} \log \frac{e^{s\left(\cos\left(\theta_{y_i}+m\right)\right)}}{e^{s\left(\cos\left(\theta_{y_i}+m\right)\right)}+ \sum_{j \in D_{y_i}, j \ne y_i}e^{s\left(\cos\left(\theta_{j}+m\right)\right)}+ \sum_{k=1, k\not\in D_{y_i}}^{n}e^{s\cos\theta_k}}.
	$}
\end{equation}

\noindent The DAML $L_d$ can be simplified as follows:

\begin{equation}
\scalebox{0.95}[1]{$
	L_d = -\frac{1}{N} \sum_{i=1}^{N} \log \frac{e^{s\left(\cos\left(\theta_{y_i}+m\right)\right)}}{\sum_{j \in D_{y_i}}e^{s\left(\cos\left(\theta_{j}+m\right)\right)}+ \sum_{k=1, k\not\in D_{y_i}}^{n}e^{s\cos\theta_k}}.
	$}
\end{equation}

\subsection{Maximum Angular Loss (MAL)} \label{sec:mal}
The DAML $L_d$ enforces the inter-class discrepancy by maintaining a margin between classes of different subjects. However, the classes that correspond to the same subject are needed to be ``close'' to each other in the hypersphere. Therefore, the MAL is proposed to enforce the similarity between classes that correspond to the same subject.

As depicted in Fig. \ref{fig:dblface}, the relationship between class representations that correspond to the same subject can be represented by the angle $\theta$ between them. Ideally, the angle $\theta$ should be 0, so all class representations that correspond to the same subject can be merged into one class. However, strictly minimizing $\theta$ may lead to overfitting due to the dominance of VIS images in the training set. Therefore, instead of minimizing the angle $\theta$, the angle $\theta$ is controlled to be smaller than a maximum angle $p$. Specifically, the MAL is defined based on Hinge loss \cite{Rosasco2004} as follows:
\begin{equation}
L_m = \frac{1}{\pi n d}\sum_{i=1}^{n}\sum_{j\in D_{y_i}}\left[\arccos\left(W_i^TW_j\right)-p\right]_+,
\end{equation}
where $d = \left|D_{y_i}\right|$ is the number of classes that share the same subject with the class $i$, and $p$ is the maximum angle between class representations. The MAL $L_m$ ensures that the angle between two class representations that correspond to the same subject is smaller than the angle $p$. The angle $p$ should be significantly smaller than the margin $m$ to ensure that the angular distance between two class representations that correspond to the same subject is smaller than the angular distance between two class representations of different subjects.

Based on the two proposed loss functions, the learning objective of \ppname is defined as:
\begin{equation}
L = L_d + \alpha L_m,
\end{equation}
where $\alpha$ is the weight to balance the two loss functions $L_d$ and $L_m$.
\section{Datasets} \label{sec:datasets}
The four datasets used for training and testing are summarized in Table \ref{tab:datasets}.

\begin{table}[h]
	\centering
	\caption{Face Datasets.}
	\label{tab:datasets}
	\resizebox{\columnwidth}{!}{%
		\begin{tabular}{lrrl}
			\hline
			Dataset           & \# Subjects & \# Images & Domain                 \\ \hline
			MS-Celeb-1M       & 86,876     & 3,923,399 & VIS                \\
			Asian-Celeb       & 93,979     & 2,830,146 & VIS               \\
			CASIA NIR-VIS 2.0 & 725        & 17,580     & NIR-VIS \\
			EDGE20            & 197        & 3,241      & NIR-VIS                 \\ \hline
		\end{tabular}%
	}
\end{table}

MS-Celeb-1M \cite{Guo2016} is the largest face dataset of VIS images, which has 10M images of 100k identities. Because MS-Celeb-1M contains many noisy labels, the cleaned version of MS-Celeb-1M that provided by DeepGlint \cite{DeepGlint2018} is used for training. The cleaned MS-Celeb-1M contains about 4M aligned images of 86,876 identities. Besides the MS-Celeb-1M dataset, the Asian-Celeb \cite{DeepGlint2018} is also used for training. The Asian-Celeb dataset contains 93,979 identities with roundly 3M VIS images. The dataset that merged MS-Celeb-1M and Asian-Celeb is called Glint \cite{DeepGlint2018}.

Since Glint contains only VIS images, it is merged with the training set of CASIA NIR-VIS 2.0 for both VIS and NIR images. As mentioned above, the CASIA NIR-VIS 2.0 dataset contains 17,580 images of 725 different subjects captured in an indoor environment. Some of the subjects in the CASIA NIR-VIS 2.0 dataset wore glasses that acted as a form of facial occlusion. We follow the \textit{view1} protocol provided by the dataset to split training and testing sets. The subjects in the training and testing sets do not overlap, and the number of subjects in the training and testing sets is almost equal. Since the CASIA NIR-VIS 2.0 dataset contains only frontal face images, we only evaluate the proposed method with frontal face images.

\begin{figure}[h]
	\begin{center}
		\begin{subfigure}[b]{0.45\columnwidth}
			\includegraphics[width=\columnwidth]{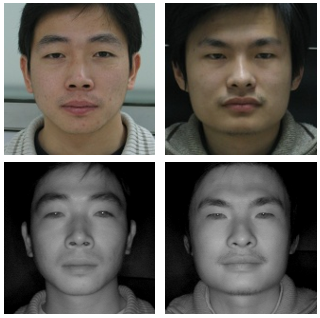}
			\caption{}
			\label{fig:casia-sample}
		\end{subfigure} \hfill
		\begin{subfigure}[b]{0.45\columnwidth}
			\includegraphics[width=\columnwidth]{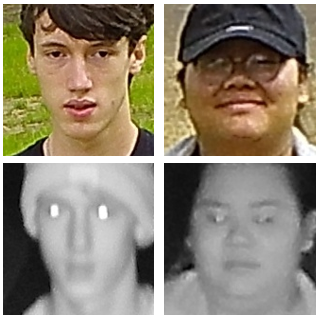}
			\caption{}
			\label{fig:edge20-sample}
		\end{subfigure}
	\end{center}
	\vspace{-0.3cm}
	\caption{Depiction of images from two testing datasets: (a) CASIA NIR-VIS 2.0 and (b) EDGE20.}
	\label{fig:casiaedge20samples}
\end{figure}

\begin{figure*}[t]
	\begin{center}
		\begin{subfigure}[b]{0.45\textwidth}
			\includegraphics[width=\textwidth]{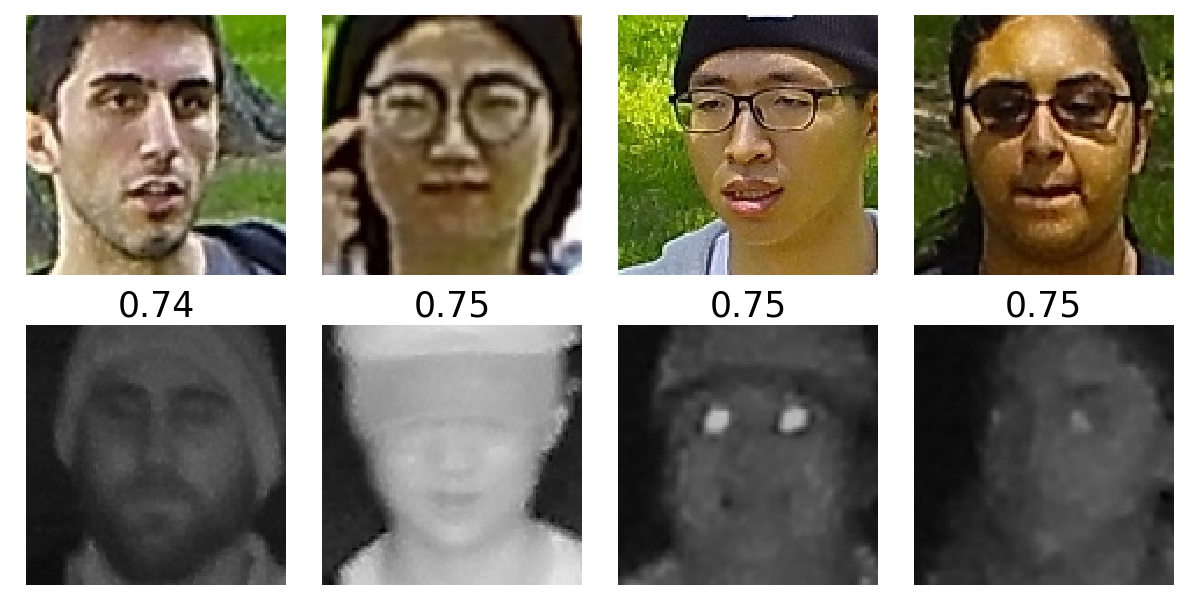}
			\caption{}
			\label{fig:edge-false1}
		\end{subfigure} \hspace{5mm}
		\begin{subfigure}[b]{0.45\textwidth}
			\includegraphics[width=\textwidth]{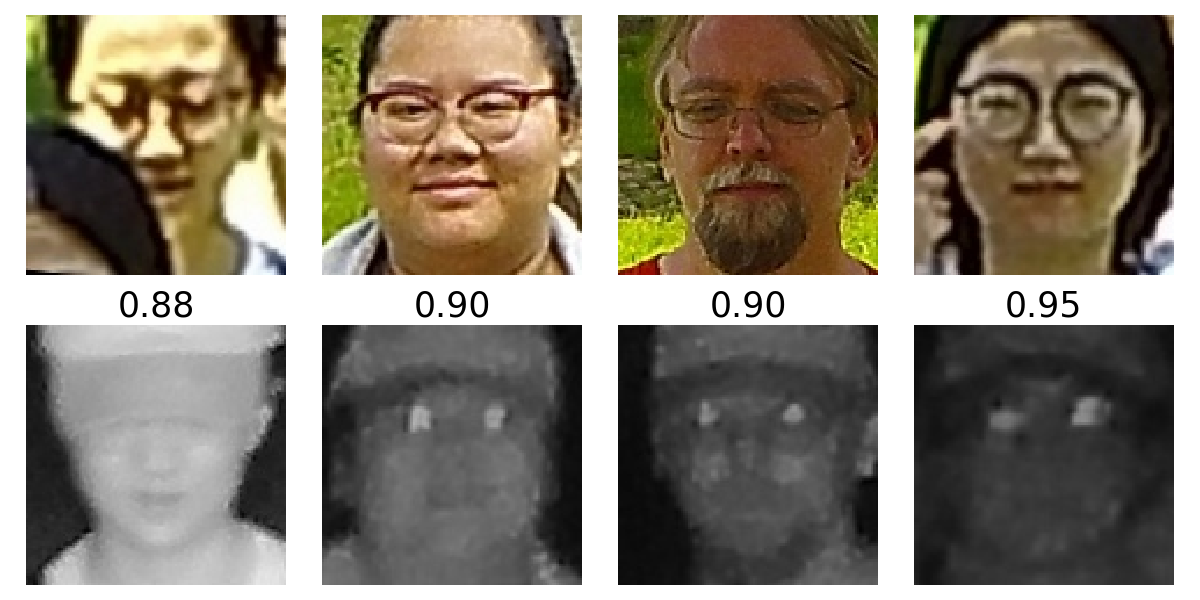}
			\caption{}
			\label{fig:edge-false2}
		\end{subfigure}
	\end{center}
	\vspace{-5mm}
	\caption{Depiction of failed matching pairs of \ppname on the EDGE20 dataset: (a) false-positive pairs and (b) false-negative pairs. The images in the first row are from the gallery, and the images in the second row are from the probes. The number in between each pair is the cosine similarity.}
	\label{fig:edge-false}
	\vspace{-3mm}
\end{figure*}

Since the CASIA NIR-VIS 2.0 dataset is acquired in a lab-controlled environment, it is no longer a challenging dataset. Thus, in addition to CASIA NIR-VIS 2.0, the EDGE20 dataset \cite{Le2020} is used for testing. The images of EDGE20 are captured from trail cameras during both day and night in an unconstrained environment. EDGE20 contains 3,241 images of 197 subjects, in which 2,500 face images in the VIS spectrum and 741 images in the NIR spectrum. Figure \ref{fig:casiaedge20samples} depicts some pairs of VIS and NIR images from the CASIA NIR-VIS 2.0 and the EDGE20 datasets. The EDGE20 dataset is more challenging than CASIA NIR-VIS 2.0 due to occlusion, low-resolution, and motion blur.
We follow the evaluation protocol \textit{FR3} defined by EDGE20. The protocol \textit{FR3} evaluates the performance of a NIR-VIS face recognition model using a set of VIS and frontal faces as gallery and using all available NIR and frontal faces as probe. The numbers of images of the gallery and the probe are 68 and 164, respectively. Each image in the gallery represents a different subject.

\section{Implementation} \label{sec:implementation}
All models in our approach are implemented using MXNet \cite{Chen2015}. The Dlib DNN face detector \cite{King2009} is used to detect faces, and PRNet \cite{Feng2018a} is used for landmark detection. Due to the limited performance of PRNet on NIR images, we manually annotated landmarks for the EDGE20 dataset to alleviate the error of misalignment.
All images are cropped and aligned to a size of $112\times112$ and normalized to the range $\left[-1, 1\right] $. We follow the same network architecture used in ArcFace, which is Resnet101. The angular margin $m$ is set to 0.5, which yields the best performance on ArcFace. The maximum angle $p$ is set to 0.15 based on our parameter tuning. During training, the batch size is set as 256, and models are trained using four GPUs. The SGD optimizer is employed with the learning rate set to 0.1 and the weight decay set to $5e-4$. The learning rate is started from 0.1 and divided by ten at 8, 10, and 12 epochs. Based on our experiments, we notice that the MAL $L_m$ converges faster than the DAML $L_d$. Therefore, the weight $\alpha$ is set to 0.5 to balance the convergence speed between the two losses. During testing, the last fully connected layer is removed. The output 512-dimensional feature embedding is used as a face template. Cosine distance is used to measure the similarity between two face templates.

In an ideal world, a cross-spectral face dataset would contain face images with uniform distribution of the number of images in each domain. In this case, a batch of training images can be equally sampled from each visual domain without repetition. However, this is not the case with existing face datasets since the number of NIR images is far fewer than the number of VIS images. A simple strategy for DBS is to equally sample data from each domain so that a batch of images contains the same number of images from both visual domains. However, the equally sampling strategy prolongs the training time due to the repeated training on the same NIR images in an epoch. Therefore, for each batch of images, we sample more images in the VIS domain to maintain a balance between the training time and the balance of training data.
In particular, for a batch of 256 images, 192 VIS images (75\%) are sampled from the the Glint dataset, 32 VIS images (12.5\%) are sampled from the VIS images of the CASIA NIR-VIS 2.0 training set, and 32 NIR images (12.5\%) are sampled from the NIR images of the CASIA NIR-VIS 2.0 training set.

\section{Experiments} \label{sec:experiments}
\subsection{Results}
In this section, we evaluate the performance of \ppname on EDGE20 and CASIA NIR-VIS 2.0 by comparing the identification rate and the verification rate with the state-of-the-art methods on NIR-VIS face recognition.

\textbf{EDGE20.} The EDGE20 dataset is a challenging dataset due to the poor quality of the images. Figure \ref{fig:edge-false} depicts the failed matching pairs of \ppname on EDGE20. These pairs clarify the challenges of EDGE20, including low resolution, motion blur, and occlusion. Therefore, a model trained barely on CASIA NIR-VIS 2.0 tends to be overfitted and does not generalize well to EDGE20.

\begin{figure}[h]
	\vspace{-3mm}
	\begin{center}
		\includegraphics[width=1.0\linewidth]{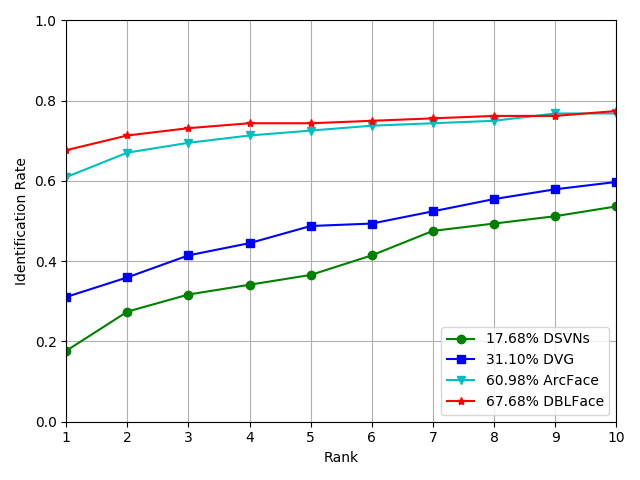}
	\end{center}
 	\vspace{-5mm}
	\caption{CMC of \ppname in comparison with DSVNs, DVG, and ArcFace on EDGE20.}
	\label{fig:cmc-dblface-comparison}
\end{figure}

Table \ref{tab:comparison-edge} presents a comparison of DBLFace with the state-of-the-art NIR-VIS face recognition methods on the EDGE20 dataset, and Fig. \ref{fig:cmc-dblface-comparison} is the corresponding CMC curve. The code and pretrained model of DSVNs is shared publicly by the authors. The simplified version of DVG is shared publicly by the author. We followed the instructions of DVG and trained a model for evaluation.
Since DSVNs is finetuned on the training set of CASIA NIR-VIS 2.0, it is overfitted and shows poor performance on EDGE20. DVG is trained on a synthetic dataset that derived from the CASIA NIR-VIS 2.0 dataset. Therefore, the DVG model performs better than the DSVNs model but does not generalize well to the images of EDGE20. Even though the baseline ArcFace is not designed for cross-spectral matching, it still outperforms both DSVNs and DVG.
The great performance of ArcFace is based on the additive angular margin loss and the large-scale training dataset Glint. The proposed method \ppname extended the additive angular margin loss by the two loss functions DAML and MAL. The two loss functions allow \ppname to be trained on the joint dataset of Glint and CASIA NIR-VIS 2.0. Thus, \ppname significantly enhanced the rank-1 identification rate of the baseline model ArcFace by 6.7\%. The verification rate at the FAR=0.1\% is also enhanced by 2.9\%.

\begin{table}[h]
	\centering
	\caption{Comparison of \ppname with the state-of-the-art NIR-VIS face recognition methods on the EDGE20 dataset. The numbers under the Rank-1 column are the identification rate, and the numbers under the VR@FAR=0.1\% column are the verification rate at the false acceptance rate of 0.1\%. The percentage sign (\%) after each number is omitted for short.}
	\label{tab:comparison-edge}
	\begin{tabular}{lcc}
		\hline
		Method          & Rank-1 & VR@FAR=0.1\% \\ \hline
		DSVNs \cite{Hu2019} & $17.68$ & $26.97$ \\
		DVG \cite{Fu2019} & $31.10$ & $39.78$ \\
		ArcFace \cite{Deng20181} &  $60.98$    &     $62.94$             \\		\hline	
		\ppname &     $\textbf{67.68}$          &       $\textbf{65.84}$     \\			\hline
	\end{tabular}%
\end{table}

McNemar's test \cite{McNemar1947} is conducted to verify if the improvement of \ppname on EDGE20 is statistically significant. Table \ref{tab:edge_contingency} is the contingency table based on rank-1 identification performance of ArcFace and \ppname.
McNemar's test with continuity correction gives the chi-square statistic value $\chi^2=4.35$ and the corresponding p-value $p=0.038$. The p-value is below the common significance level 0.05, which confirms that the performance difference between ArcFace and \ppname is statistically significant.

\begin{table}[h]
	\centering
	\caption{Contingency table based on rank-1 identification performance of ArcFace and \ppname.}
	\label{tab:edge_contingency}
	\begin{tabular}{ccc}
		\hline
		& \multicolumn{2}{c}{\ppname} \\ \cline{2-3} 
		ArcFace & + & - \\ \hline
		+ & 94 & ~6 \\
		- & 17 & 47 \\ \hline
	\end{tabular}%
	
\end{table}

\begin{table}[h]
	\centering
	\caption{Comparison of \ppname with the state-of-the-art NIR-VIS face recognition methods on the CASIA NIR-VIS 2.0 dataset. The numbers under the Rank-1 column are the identification rate, and the numbers under the VR@FAR=0.1\% column are the verification rate at the false acceptance rate of 0.1\%. The percentage sign (\%) after each number is omitted for short.}
	\label{tab:comparison-casia}
	\begin{tabular}{lcc}
		\hline
		Method          & Rank-1 & VR@FAR=0.1\% \\ \hline
		DSVNs \cite{Hu2019} & $98.20$ & $97.30$ \\
		CFC \cite{He2020} & $99.21$ & $98.81$ \\
		RGM \cite{Cho2020} & $99.30$ & $98.90$ \\
		RCN \cite{Deng2019}  & $99.32$ & $98.74$ \\
		DVG \cite{Fu2019} & $99.80$ & $99.80$ \\
		MMDL \cite{Cao2019a} & $99.90$ & $99.40$ \\
		ArcFace \cite{Deng20181} &  $99.97$    &     $99.96$             \\		\hline	
		\ppname &     $\textbf{99.98}$          &       $\textbf{99.97}$     \\			\hline
	\end{tabular}%
\end{table}

\begin{figure}[h]
	\begin{center}
		\begin{subfigure}[b]{0.35\columnwidth}
			\includegraphics[width=\columnwidth]{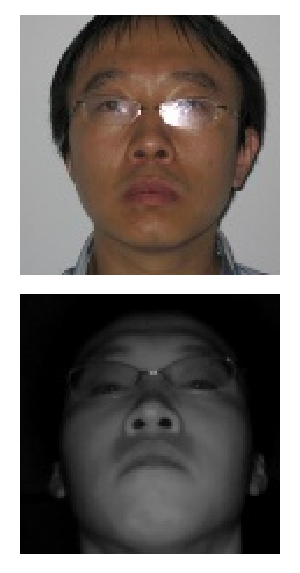}
			\caption{}
			\label{fig:casia-false1}
		\end{subfigure}
		\begin{subfigure}[b]{0.35\columnwidth}
			\includegraphics[width=\columnwidth]{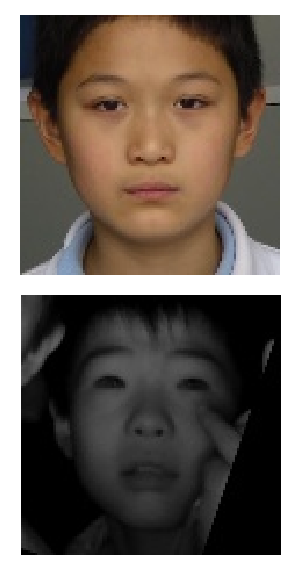}
			\caption{}
			\label{fig:casia-false2}
		\end{subfigure}
	\end{center}
		\vspace{-5mm}
	\caption{The two false-negative pairs of ArcFace and \ppname on CASIA NIR-VIS 2.0: Both (a) and (b) are false-negative pairs of ArcFace, and (b) is the only false-negative pair of \ppname.}
	\label{fig:casia-false}
\end{figure}

\textbf{CASIA NIR-VIS 2.0.} Even though the performance of the state-of-the-art methods is saturated on the CASIA NIR-VIS 2.0 dataset, it remains the most popular benchmark for NIR-VIS face recognition. Therefore, we evaluate the performance of \ppname on the CASIA NIR-VIS 2.0 dataset for comparison purposes. Table \ref{tab:comparison-casia} shows a comparison of \ppname with the state-of-the-art NIR-VIS face recognition methods on the CASIA NIR-VIS 2.0 dataset.
DSVNs and DVG achieve high performance on CASIA NIR-VIS 2.0 but perform poorly on EDGE20. The behavior of DSVNs and DVG clarifies the lack of generalization of the methods that relied on the CASIA NIR-VIS 2.0 for fine-tuning.
The baseline ArcFace already outperforms other methods on both rank-1 identification rate and verification rate at the FAR=0.1\%. \ppname demonstrates improved performance over ArcFace and achieves the best performance among state-of-the-art methods. Figure \ref{fig:casia-false} lists all the false-negative cases of ArcFace and \ppname on the CASIA NIR-VIS 2.0 dataset. ArcFace contains only two false-negative pairs, while the number of false-negative pairs of \ppname is one. The first false-negative pair of ArcFace is challenging due to the glare on glasses and the head pose. \ppname indirectly overcomes these challenges by narrowing down the domain discrepancy. The common false-negative pair of both ArcFace and \ppname is due to mislabeling. The two images belong to two different subjects. Images on CASIA NIR-VIS 2.0 are supposed to contain only one face per image. However, several images contain more than one face. Our face detector selected the first detected face, which may not the subject of interest.

\vspace{-5mm}
\subsection{Parameter Tuning}
In this section, we study the impact of the parameter $p$ to the performance of \ppname. The parameter tuning tests are conducted on a subset of the \textit{FR3} protocol of the EDGE20 dataset. In this subset, the numbers of images of the gallery and the probe are 42 and 96, respectively.

\begin{figure}[h]
	\begin{center}
		\includegraphics[width=1.0\linewidth]{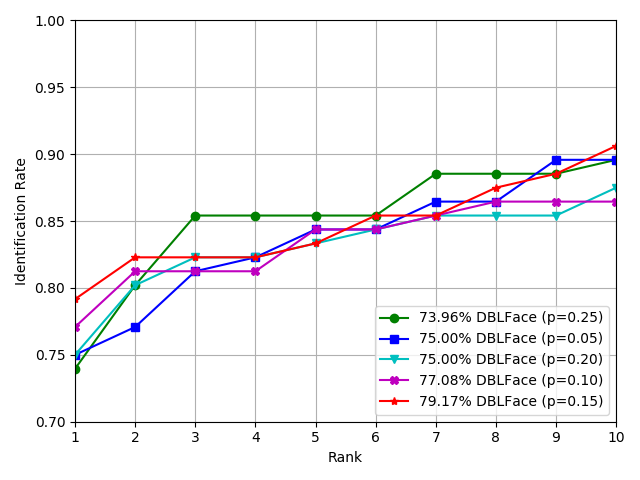}
	\end{center}
		\vspace{-5mm}
	\caption{CMC of \ppname on EDGE20.}
	\vspace{-5mm}
	\label{fig:cmc-dblface}
\end{figure}

The angle $p$ represents the maximum angle between two class representations that correspond to the same subject. The value of $p$ should be significantly smaller than the margin $m$, which is set to $0.5$. Therefore, we trained five versions of \ppname by setting the maximum angle $p$ to $0.05$, $0.1$, $0.15$, $0.2$, and $0.25$. Table \ref{tab:ablation-dblface} shows the performance of \ppname with different settings of the angle $p$, and Fig. \ref{fig:cmc-dblface} is the corresponding CMC curve. The smaller the value of $p$, the harder the enforcement to the relationship between two class representations that correspond to the same subject. We observe that the peak performance is achieved with $p=0.15$. When the value of $p$ is decreased, the model tends to be overfitted to the training data. When the value of $p$ is increased, the model loosens up the enforcement. As shown in Fig. \ref{fig:cmc-dblface}, loosening the enforcement reduces the identification rate at rank 1 and 2 but allows the model to perform well at a larger rank.

\begin{table}[h]
	\centering
	\caption{The impact of the angle $p$ to the performance of \ppname. A subset of the \textit{FR3} protocol of the EDGE20 dataset is used for evaluation. The numbers under the Rank-1 column are the identification rate, and the numbers under the VR@FAR=0.1\% column are the verification rate at the false acceptance rate of 0.1\%. The percentage sign (\%) after each number is omitted for short.}
	\label{tab:ablation-dblface}
	\begin{tabular}{cccc}
		\hline
		Angle $p$          & Rank-1          & VR@FAR=0.1\% \\ \hline
		0.05 & $75.00$        &     $67.25$                      \\		
		0.10 & $77.08$        &     $69.62$                      \\			
		0.15 & $\textbf{79.17}$   &     $\textbf{71.33}$                      \\			
		0.20 & $75.00$        &     $69.74$                      \\
		0.25 & $73.96$        &     $70.96$                      \\						 \hline
	\end{tabular}%
\end{table}

\section{Conclusion} \label{sec:conclusion}
In this paper, we described the challenges of large intra-class variation and data imbalance in NIR-VIS heterogeneous face recognition. To address these challenges, we introduced the idea that a subject is represented by two labels, one for the VIS images and one for the NIR images. Based on this idea, we proposed \ppname, a learning approach based on the DAML and the MAL. The DAML is presented to maintain the discrepancy between classes of different subjects. The MAL is presented to control the close relationship of class representations that correspond to the same subject. We evaluated \ppname with EDGE20 and CASIA NIR-VIS 2.0. Through our extensive experiments, we demonstrated that \ppname achieves statistically significant improvement on the EDGE20 dataset and attains the state-of-the-art result on the CASIA NIR-VIS 2.0 dataset.

\noindent\textbf{Acknowledgment}
This material is based upon work supported by the U.S. Department of Homeland Security under Grant Award Number 2017-ST-BTI-0001-0201. This grant is awarded to the Borders, Trade, and Immigration (BTI) Institute: A DHS Center of Excellence led by the University of Houston, and includes support for the project EDGE awarded to the University of Houston. The views and conclusions contained in this document are those of the authors and should not be interpreted as necessarily representing the official policies, either expressed or implied, of the U.S. Department of Homeland Security.

{\small
\bibliographystyle{ieee}
\bibliography{library}

\begin{thebibliography}{10}\itemsep=-1pt

\bibitem{Cao2019a}
B.~Cao, N.~Wang, X.~Gao, J.~Li, and Z.~Li.
\newblock {Multi-margin based decorrelation learning for heterogeneous face
  recognition}.
\newblock In {\em Proc. International Joint Conference on Artificial
  Intelligence}, pages 680--686, Macao, China, Aug. 2019.

\bibitem{Chen2015}
T.~Chen, M.~Li, Y.~Li, M.~Lin, N.~Wang, M.~Wang, T.~Xiao, B.~Xu, C.~Zhang, and
  Z.~Zhang.
\newblock {MXNet: A flexible and efficient machine learning library for
  heterogeneous distributed systems}.
\newblock {\em arXiv preprint}, Dec. 2015.

\bibitem{Cho2019}
M.~Cho, T.-Y. Chung, T.~Kim, and S.~Lee.
\newblock {NIR-to-VIS face recognition via embedding relations and coordinates
  of the pairwise features}.
\newblock In {\em Proc. International Conference on Biometrics}, pages 1--8,
  Crete, Greece, Jun. 2019.

\bibitem{Cho2020}
M.~Cho, T.~Kim, I.-J. Kim, and S.~Lee.
\newblock {Relational deep feature learning for heterogeneous face
  recognition}.
\newblock {\em arXiv preprint}, Mar. 2020.

\bibitem{DeepGlint2018}
DeepGlint.
\newblock {Trillionpairs (http://trillionpairs.deepglint.com)}, 2018.

\bibitem{Deng20181}
J.~Deng, J.~Guo, and S.~Zafeiriou.
\newblock {ArcFace: Additive angular margin loss for deep face recognition}.
\newblock In {\em Proc. IEEE Conference on Computer Vision and Pattern
  Recognition}, pages 1--12, Long Beach, CA, Jun. 2019.

\bibitem{Deng2019}
Z.~Deng, X.~Peng, and Y.~Qiao.
\newblock {Residual compensation networks for heterogeneous face recognition}.
\newblock In {\em Proc. AAAI Conference on Artificial Intelligence}, pages
  8239--8246, Jul. 2019.

\bibitem{Farokhi2016}
S.~Farokhi, J.~Flusser, and U.~{Ullah Sheikh}.
\newblock {Near infrared face recognition: A literature survey}.
\newblock {\em Computer Science Review}, 21:1--17, Aug. 2016.

\bibitem{Feng2018a}
Y.~Feng, F.~Wu, X.~Shao, Y.~Wang, and X.~Zhou.
\newblock {Joint 3D face reconstruction and dense alignment with position map
  regression network}.
\newblock In {\em Proc. European Conference in Computer Vision}, Munich,
  Germany, Sep. 2018.

\bibitem{Fu2019}
C.~Fu, X.~Wu, Y.~Hu, H.~Huang, and R.~He.
\newblock {Dual variational generation for low-shot heterogeneous face
  recognition}.
\newblock In {\em Proc. Advances in Neural Information Processing Systems},
  pages 1--10, Vancouver, Canada, Dec. 2019.

\bibitem{Goodfellow2014}
I.~J. Goodfellow, J.~Pouget-Abadie, M.~Mirza, B.~Xu, D.~Warde-Farley, S.~Ozair,
  A.~Courville, and Y.~Bengio.
\newblock {Generative adversarial nets}.
\newblock In {\em Proc. Advances in Neural Information Processing Systems},
  pages 2672--2680, Montreal, Canada, 2014.

\bibitem{Guillaumin2009}
M.~Guillaumin, J.~Verbeek, and C.~Schmid.
\newblock {Is that you? Metric learning approaches for face identification}.
\newblock In {\em Proc. IEEE International Conference on Computer Vision},
  pages 498--505, Kyoto, Japan, Oct. 2009.

\bibitem{Guo2016}
Y.~Guo, L.~Zhang, Y.~Hu, X.~He, and J.~Gao.
\newblock {MS-Celeb-1M: A dataset and benchmark for large-scale face
  recognition}.
\newblock In {\em Proc. European Conference on Computer Vision}, pages 87--102,
  Amsterdam, The Netherlands, Oct. 2016.

\bibitem{He2020}
R.~He, J.~Cao, L.~Song, Z.~Sun, and T.~Tan.
\newblock {Adversarial cross-spectral face completion for NIR-VIS face
  recognition}.
\newblock {\em IEEE Transactions on Pattern Analysis and Machine Intelligence},
  42(5):1025--1037, May 2020.

\bibitem{He2018}
R.~He, X.~Wu, Z.~Sun, and T.~Tan.
\newblock {Wasserstein CNN: Learning invariant features for NIR-VIS face
  recognition}.
\newblock {\em IEEE Transactions on Pattern Analysis and Machine Intelligence},
  41(7):1761 -- 1773, 2018.

\bibitem{Hu2019b}
W.~Hu and H.~Hu.
\newblock {Discriminant deep feature learning based on joint supervision loss
  and multi-layer feature fusion for heterogeneous face recognition}.
\newblock {\em Computer Vision and Image Understanding}, 184:9--21, Jul. 2019.

\bibitem{Hu2019}
W.~Hu and H.~Hu.
\newblock {Disentangled spectrum variations networks for NIR–VIS face
  recognition}.
\newblock {\em IEEE Transactions on Multimedia}, 22(5):1234--1248, May 2020.

\bibitem{JieChen2010}
{Jie Chen}, D.~Yi, {Jimei Yang}, {Guoying Zhao}, S.~Z. Li, and M.~Pietikainen.
\newblock {Learning mappings for face synthesis from near infrared to visual
  light images}.
\newblock In {\em Proc. IEEE Conference on Computer Vision and Pattern
  Recognition}, pages 156--163, Miami, FL, Jun. 2009.

\bibitem{Juefei-Xu2015a}
F.~Juefei-Xu, D.~K. Pal, and M.~Savvides.
\newblock {NIR-VIS heterogeneous face recognition via cross-spectral joint
  dictionary learning and reconstruction}.
\newblock In {\em Proc. IEEE Conference on Computer Vision and Pattern
  Recognition Workshops}, pages 141--150, Boston, MA, Oct. 2015.

\bibitem{King2009}
D.~E. King.
\newblock {Dlib-ml: A machine learning toolkit}.
\newblock {\em Journal of Machine Learning Research}, 10:1755--1758, 2009.

\bibitem{Le2020}
H.~Le, C.~Smailis, L.~Shi, and I.~A. Kakadiaris.
\newblock {EDGE20: A cross spectral evaluation dataset for multiple
  surveillance problems}.
\newblock In {\em Proc. IEEE Winter Conference on Applications of Computer
  Vision}, pages 1--10, Snowmass Village, CO, Mar. 2020.

\bibitem{Lezama2017}
J.~Lezama, Q.~Qiu, and G.~Sapiro.
\newblock {Not afraid of the dark: NIR-VIS face recognition via cross-spectral
  hallucination and low-rank embedding}.
\newblock In {\em Proc. IEEE Conference on Computer Vision and Pattern
  Recognition}, pages 6807--6816, Honolulu, HI, Jul. 2017.

\bibitem{Li2013}
S.~Z. Li, D.~Yi, Z.~Lei, and S.~Liao.
\newblock {The CASIA NIR-VIS 2.0 face database}.
\newblock In {\em Proc. IEEE Conference on Computer Vision and Pattern
  Recognition Workshops}, pages 348--353, Portland, OR, Jun. 2013.

\bibitem{Liu2017}
W.~Liu, Y.~Wen, Z.~Yu, M.~Li, B.~Raj, and L.~Song.
\newblock {SphereFace: Deep hypersphere embedding for face recognition}.
\newblock In {\em Proc. IEEE Conference on Computer Vision and Pattern
  Recognition}, pages 6738--6746, Honolulu, HI, Jul. 2017.

\bibitem{Liu2016b}
X.~Liu, L.~Song, X.~Wu, and T.~Tan.
\newblock {Transferring deep representation for NIR-VIS heterogeneous face
  recognition}.
\newblock In {\em Proc. International Conference on Biometrics}, Halmstad,
  Sweden, Aug. 2016.

\bibitem{McNemar1947}
Q.~McNemar.
\newblock {Note on the sampling error of the difference between correlated
  proportions or percentages}.
\newblock {\em Psychometrika}, 12(2):153--157, Jun. 1947.

\bibitem{Ouyang2016}
S.~Ouyang, T.~Hospedales, Y.~Z. Song, X.~Li, C.~C. Loy, and X.~Wang.
\newblock {A survey on heterogeneous face recognition: Sketch, infra-red, 3D
  and low-resolution}.
\newblock {\em Image and Vision Computing}, 56:28--48, Dec. 2016.

\bibitem{Peng2019}
C.~Peng, N.~Wang, J.~Li, and X.~Gao.
\newblock {Re-ranking high-dimensional deep local representation for NIR-VIS
  face recognition}.
\newblock {\em IEEE Transactions on Image Processing}, 28(9):4553--4565, Sep.
  2019.

\bibitem{Rosasco2004}
L.~Rosasco, E.~{De Vito}, A.~Caponnetto, M.~Piana, and A.~Verri.
\newblock {Are loss functions all the same?}
\newblock {\em Neural Computation}, 16(5):1063--1076, May 2004.

\bibitem{Saxena2016}
S.~Saxena and J.~Verbeek.
\newblock {Heterogeneous face recognition with CNNs}.
\newblock In {\em Proc. European Conference in Computer Vision Workshops},
  pages 483--491, Amsterdam, The Netherlands, Oct. 2016.

\bibitem{Schroff2015}
F.~Schroff, D.~Kalenichenko, and J.~Philbin.
\newblock {FaceNet: A unified embedding for face recognition and clustering}.
\newblock In {\em Proc. IEEE Conference on Computer Vision and Pattern
  Recognition}, pages 815--823, Boston, MA, Jun. 2015.

\bibitem{Song2018}
L.~Song, M.~Zhang, X.~Wu, and R.~He.
\newblock {Adversarial discriminative heterogeneous face recognition}.
\newblock In {\em Proc. AAAI Conference on Artificial Intelligence}, New
  Orleans, LA, Feb. 2018.

\bibitem{Wu2018}
X.~Wu, L.~Song, R.~He, and T.~Tan.
\newblock {Coupled deep learning for heterogeneous face recognition}.
\newblock In {\em Proc. AAAI Conference on Artificial Intelligence}, pages
  1679--1686, New Orleans, LA, Feb. 2018.

\end{thebibliography}
}

\end{document}